\documentclass{article}

\usepackage[position, preprint]{neurips_2026}

\usepackage[utf8]{inputenc}
\usepackage[T1]{fontenc}
\usepackage{hyperref}
\hypersetup{
    colorlinks,
    linkcolor={red!50!black},
    citecolor={blue!50!black},
    urlcolor={blue!80!black}
}
\usepackage{url}
\usepackage{comment}
\usepackage{booktabs}
\usepackage{amsfonts}
\usepackage{amsmath}
\usepackage{nicefrac}
\usepackage{microtype}
\usepackage{graphicx}
\usepackage{caption}
\usepackage{subcaption}
\usepackage[dvipsnames]{xcolor}
\usepackage{cprotect}
\usepackage{bm}
\usepackage{wrapfig}
\usepackage{mathtools}
\usepackage{enumitem}   
\usepackage{wrapfig}
\usepackage{float}
\usepackage{url}
\usepackage{upgreek}
\usepackage{tikz-cd}

\title{Understanding diffusion models \\
requires rethinking (again) generalization}

\author{%
  Pierre Marion\thanks{Equal Contribution}\\
  Inria, \'Ecole Normale Sup\'erieure, PSL Research University\\
  \texttt{pierre.marion@inria.fr}\\
  \And
  Yu-Han Wu\footnotemark[1]\\
  LPSM, Sorbonne University \& Google DeepMind\\
  \texttt{yhwu@google.com}
}

\begin{document}

\maketitle
\begin{abstract}
This position paper argues that understanding generalization in diffusion models requires fundamentally new theoretical frameworks that go beyond both classical statistical learning theory and the benign overfitting paradigm developed for supervised learning.
In diffusion models, unlike in supervised learning, memorization of training data and generalization to novel samples are incompatible: a model that has fully memorized its training set generates copies rather than novel data.
Several theoretical explanations for why practical diffusion models nevertheless generalize have been proposed, based on capacity limitations, implicit regularization from optimization, or architectural inductive biases, but their interactions remain unclear. We argue that the field should pivot from explaining why the diffusion models do not memorize to investigating what the model actually learns during pre-memorization phase. To highlight our stance, we conduct empirical study of diffusion models trained on CIFAR-10, and we distill the findings into concrete open questions that we believe are key to improve understanding of generalization in diffusion models.

\end{abstract}

\section{Introduction}
\label{sec:intro}

Classical statistical learning theory provides the foundational tools for reasoning about generalization in machine learning.
Through notions such as VC dimension and Rademacher complexity~\citep{bartlett2002rademacher,shalev2014understanding}, it establish that the generalization gap scales with the model's capacity to fit arbitrary labels. %
This proved remarkably successful for  traditional methods~\citep{bousquet2002stability}, but \citet{arpit2017closer,neyshabur2017exploring,zhang2017understanding} showed that modern deep neural networks can perfectly fit randomly labeled data, yet generalize well on true labels, rendering classical bounds vacuous for practical deep learning and prompted a fundamental rethinking of generalization.

The past decade has witnessed an intense effort to bridge this gap.
The theory of benign overfitting~\citep{bartlett2020benign} and double descent~\citep{belkin2019reconciling,nakkiran2019deep} showed that overparameterized models can simultaneously interpolate the training data and generalize well to unseen examples.
This is closely connected to the study of implicit regularization by the optimizer~\citep{gunasekar2017implicit,chizat2020implicit}.
A complementary perspective comes from the PAC-Bayes framework~\citep{mcallester1999pac}, which has been used to derive non-vacuous generalization bounds for deep neural networks~\citep{dziugaite2017computing}; see~\citet{alquier2024user} for a recent survey.

The situation in generative models, and in particular diffusion models~\citep{ho2020denoising,song2021score}, is fundamentally different.
Diffusion models are trained by denoising score matching \citep{hyvarinen2005estimation, vincent2011connection}, in which the empirical risk is an expectation over Gaussian noise added to each training sample.
Because this noise is continuous, the training loss cannot be driven to zero; instead, the global minimum of the empirical score matching loss corresponds to the score function of the noised empirical measure, i.e., a mixture of Gaussians centered at the training points~\citep{li2024good}. Sampling from it produces near-exact copies of the training data~\citep{somepalli2023diffusion,somepalli2023understanding,carlini2023extracting,gu2023memorization,yoon2023diffusion}.
In other words, unlike in supervised learning where interpolation and generalization can coexist, reaching the global minimum of the empirical risk in denoising score matching is fundamentally detrimental.
Yet practical diffusion models clearly generate diverse, novel samples, implying that some form of regularization prevents full memorization.
This question is important not only from a theoretical standpoint, but crucial also in practice, particularly regarding privacy risks and intellectual property~\citep{carlini2023extracting,vyas2023provable,zhang2023copyright}.

Several concurrent theoretical explanations have been proposed for this regularization, falling broadly into three families:
(i)~capacity limitations of the network architecture relative to the dataset size~\citep{gu2023memorization,yoon2023diffusion,buchanan2025edge},
(ii)~implicit regularization from the optimization dynamics~\citep{bonnaire2025diffusion,favero2025bigger,wu2025taking}, and
(iii)~inductive biases of the architecture~\citep{kadkhodaie2023generalization,kamb2024analytic,cui2024analysis}.
Each captures an important facet of the phenomenon, but their interactions remain poorly understood.

This motivates three research goals:
\begin{enumerate}[topsep=0pt,itemsep=2pt,parsep=3pt,leftmargin=30pt]
    \item a finer understanding of how the various sources of regularization interact;
    \item an assessment of the match between theoretical predictions and empirical observations;
    \item a delineation between well-understood phenomena and interesting open questions.
\end{enumerate}

Before stating our position, we clarify two distinct aspects of generative model behavior that are sometimes conflated under the single term ``generalization.''
We use \emph{novelty} to refer to the model's ability to produce samples that are not copies of training data (i.e., the absence of memorization), and \emph{fidelity} to refer to the quality and distributional accuracy of the generated samples with respect to the true data distribution.
A model that generates pure noise has perfect novelty but zero fidelity; a model that copies training examples has high fidelity but no novelty.
What we informally call \emph{generalization} in diffusion models is the simultaneous achievement of both novelty and fidelity.

\textbf{Position.}
\textbf{We argue that understanding generalization in diffusion models requires going beyond both classical statistical learning theory and the benign overfitting paradigm that reshaped our understanding of supervised learning.
We contend that the question of \emph{why} diffusion models do not memorize in large-scale applications is largely resolved: early stopping, combined with the linear scaling of memorization time with dataset size.
The deeper and most pressing question in the theoretical analysis of diffusion models is what is actually learned during the pre-memorization phase, and how novelty and fidelity jointly emerge from the interplay between optimization, architecture, and data geometry.}

\paragraph{Methodology.}
We perform empirical studies on CIFAR-10~\citep{krizhevsky2009learning} with U-Net architectures~\citep{ronneberger2015u} to support our proposed view. We conduct careful hyperparameter sweeps over dataset size, model size, batch size, and learning rate while tracking multiple metrics throughout training.
This intermediate-scale study is positioned between the Gaussian data of theoretical analyses and the large-scale experiments of practical systems.
A limitation is that it is not entirely clear whether all findings scale to larger models and datasets; we revisit this question later.
Nevertheless, we argue that the observations provide sufficient evidence to inform important research directions.

While most of our observations connect to existing theoretical predictions, some phenomena (for instance the role of the batch size) appear to be documented for the first time. Further, systematic metrics reporting across a unified experimental setup had not been present in papers providing theoretical analysis of memorization in diffusion models. We emphasize that the purpose of these experiments is not to analyze new phenomena or methodology, but rather to test the match between existing theoretical predictions and empirical observations. Informed by these findings, we outline our perspective on the state of the field and identify concrete open questions.

\paragraph{Alternative views.} Our position differs from three prevailing directions in the theoretical analysis of diffusion models. The first one aims at bounding the error propagation through the backward diffusion, \textit{assuming} that the true score has been learned up to small error \citep{benton2024nearly,beyler2025convergence,conforti2025kl,strasman2025an}. While this has been an important contribution, we argue that error propagation is now well-understood, including fine-grained questions such as the interaction between stochasticity and discretization step, and that we now need to focus on the learning problem. The second view is to explain generalization of diffusion models through proving convergence rates in distributional distance \citep[e.g.,][]{oko2023diffusion,azangulov2024convergence,tang2024adaptivity,lyu2025resolving,stephanovitch2025generalization}. This would show that diffusion models do more than memorizing the training data since the empirical distribution does not match their convergence rate.
This analysis is typically performed through capacity arguments, by considering the score minimizing the empirical risk.  
Our experimental assessment suggests that this approach is unlikely to explain the practical success of diffusion models for two reasons: the train-test gap in empirical distributional distance (e.g., the difference between Train FID and Test FID) does not directly measure fidelity to the true data distribution, and neural networks used in standard pipelines do have the capacity to memorize the training data. The third alternative position is to study \emph{why} diffusion models do not memorize \citep{bonnaire2025diffusion,buchanan2025edge,favero2025bigger,wu2025taking,farghly2026implicit}. We argue that this question is essentially solved and that the field should move forward to understanding \textit{what is learned} by the model.
\section{Current explanations of memorization in diffusion models}
\label{sec:background}

In this section we survey the current literature on memorization in diffusion models, organizing the proposed explanations into three families.
The central empirical observation motivating this line of work is that when a diffusion model is trained long enough on a sufficiently small dataset, it transitions from generating novel samples to reproducing training examples~\citep{gu2023memorization,yoon2023diffusion,somepalli2023diffusion,carlini2023extracting}.

\subsection{Capacity of the network versus sample size}
\label{subsec:capacity}

A natural first explanation is that memorization occurs when the model has sufficient capacity to fit the training data.
\citet{yoon2023diffusion,kadkhodaie2023generalization,gu2023memorization} showed that when the training set is small enough the model memorizes, else it generalizes.
More recently, \citet{buchanan2025edge} studied the memorization capacity needed for the case of Gaussian mixtures.
The dependence of memorization on the ratio of model capacity to dataset size naturally echoes the classical generalization toolbox in statistical learning theory.
This connection has been made explicit in a line of theoretical works providing convergence rates for score-based generative models \citep[e.g.,][]{oko2023diffusion,azangulov2024convergence,tang2024adaptivity,lyu2025resolving,stephanovitch2025generalization}.
These works establish the statistical consistency of diffusion models but typically assume exact or near-exact minimization of the empirical risk, leaving open the question of how the interplay between finite dataset size and optimization dynamics affects generalization.

\subsection{Implicit regularization from the optimization algorithm}
\label{subsec:implicit-reg}

The training algorithm itself has also been found to play a role in preventing memorization.

\paragraph{Early stopping.}
\citet{bonnaire2025diffusion} and \citet{favero2025bigger} independently showed that early stopping acts as a crucial implicit regularizer in diffusion model training.
Both works demonstrate that the training step $\tau_{\mathrm{mem}}$ at which a diffusion model starts to memorize scales \emph{linearly} with the dataset size~$N$.
This implies that for large datasets the memorization time can be astronomically large, providing a convincing explanation of why practical models do not exhibit strong memorization.

\paragraph{Learning rate.}
\citet{wu2025taking} showed that large learning rates in gradient descent applied to the denoising score matching objective prevent memorization.
Their analysis builds on the concept of \emph{minimum stability}~\citep{mulayoff2020unique,qiao2024stable}: gradient descent with learning rate~$\eta$ can only converge to minima whose sharpness (measured by the largest eigenvalue of the Hessian) is bounded by~$O(1/\eta)$.
Since the global minimum of the score matching loss is typically sharp, a sufficiently large learning rate prevents convergence to this minimum.
This connects to the broader literature on the \emph{edge of stability}~\citep{cohen2021gradient}, where gradient descent operates near the boundary of divergence.
An important nuance is that the minimum stability argument characterizes the properties of the solution at the end of training, and not along the optimization trajectory.
This distinction will be relevant in light of our experimental findings in Section~\ref{subsec:predictions-vs-reality}.

\paragraph{Inductive bias from the score matching loss.}
A third line of work argues that the score matching objective itself contains an implicit bias toward generalization.
\citet{farghly2025diffusion} showed that the log-domain smoothing induced by score matching is geometry-adaptive, naturally favoring solutions that capture the manifold structure of the data rather than individual data points.
\citet{li2025scores} and \citet{shen2026manifold} proved rate separations demonstrating that learning through score matching the support of a low-dimensional manifold is easier than learning the density on the manifold, providing a statistical argument for why diffusion models learn the manifold structure before starting to memorize.

\subsection{Inductive bias of the architecture}
\label{subsec:architecture}

Finally, the architecture of the score network may also introduce biases that favor generalization.
\citet{kadkhodaie2023generalization} argued that convolutional architectures naturally learn geometry-adaptive harmonic representations.
\citet{kamb2024analytic} suggest that invariances imposed by architectural choices (such as convolutional weight-sharing and locality) determine the model's ability to generate novel samples.
\citet{cui2024analysis} provided a theoretical analysis of inductive biases induced by the choice of architecture in a simple flow-based generative modeling framework.

These architectural considerations are complementary to the optimization-based explanations: the architecture determines the function class, while the optimizer selects a particular function within that class.
In this paper, we do not investigate the effect of architecture because comparing different architectures in a controlled manner is delicate.
We therefore fix the architecture to U-Nets and focus on the roles of dataset size, model size, and optimization hyperparameters.

\section{Empirical study of the memorization transition}
\label{sec:experiments}

In this section, we present empirical results from training diffusion models on CIFAR-10. Our goal is to investigate how model capacity, dataset size, batch size, and learning rate influence the training dynamics and in particular the memorization in diffusion models.
The reason why we consider a relatively small scale dataset is that we discovered it is nontrivial to find an experimental setup that exhibits \emph{both} reasonable fidelity \emph{and} eventual memorization within a tractable training budget.
Obtaining memorization alone is easy: simply provide very few samples.
Achieving generalization is also straightforward with off-the-shelf training pipelines on large datasets, but then memorization is not observed.
We conjecture that for most practical setups the dataset is large enough that the time required to transition to memorization exceeds any reasonable training budget, since the theory predicts that this time scales linearly with dataset size~\citep{bonnaire2025diffusion,favero2025bigger}.

\subsection{Experimental setup}
\label{subsec:setup}

\paragraph{Model and data.} For training, we select the first $N$ samples of CIFAR-10~\citep{krizhevsky2009learning} from the training set, varying $N$ across experiments.
For evaluation, we use a fixed set of $5{,}120$ samples from the test set.
We use U-Net architectures~\citep{ronneberger2015u} with a varying number of channels ($128$ base channels by default), embedding the class-conditioning label and noise level into $512$-dimensional vectors.
We adopt the rectified flow formulation of diffusion models~\citep{liu2023flow} with $250$ sampling steps and classifier-free guidance with strength~$0.0$.

\paragraph{Metrics.}
We track the following quantities during training, computed on train and test splits:
(i)~\textbf{Denoising score matching loss}: computed on a single batch for the training set and on the full set for the test set;
(ii)~\textbf{Memorization score}~\citep{pizzi2022self,somepalli2023understanding}: the ratio of generated samples with maximum cosine similarity to the training data exceeding $0.6$ in a self-supervised feature space (SSCD~\citep{pizzi2022self});
(iii)~\textbf{Sliced Wasserstein distance}: computed between 2{,}000 generated samples and the reference split (train or test) in the Inception feature spaces;
(iv)~\textbf{FID}~\citep{heusel2017gans}: computed between generated samples and the reference split, with the number of generated samples matching the reference set size.

\paragraph{Hyperparameter sweeps.}
We vary the following hyperparameters (default values in \textbf{bold}):
\begin{itemize}[topsep=0pt,itemsep=2pt,parsep=2pt,leftmargin=10pt]
    \item[-] Dataset size $N \in \{\mathbf{2048}, 4096, 8192, 16384\}$.
    \item[-] Model size (number of parameters) $P \in \{3\text{M}, 8\text{M}, 26\text{M}, \textbf{93\text{M}}, 201\text{M}\}$.
    \item[-] Batch size $B \in \{\mathbf{16}, 32, 64, 128\}$.
    \item[-] Learning rate $\eta \in \{10^{-5},\; 5 \cdot 10^{-5},\; \mathbf{10^{-4}},\; 4 \cdot 10^{-4}\}$.
\end{itemize}

\paragraph{Quantities of interest.}
For each configuration, we examine:
(i)~the memorization time~$\tau_{\mathrm{mem}}$. This is defined as the time when the test loss begins diverging from the training loss, which is known to be associated with the onset of memorization~\citep{bonnaire2025diffusion,favero2025bigger}.
We stress that $\tau < \tau_{\mathrm{mem}}$ implies novelty (generated samples are not copies of training data) but does not automatically imply high fidelity;
(ii)~the \emph{memorization rate}, i.e., the slope of the memorization score once in the memorization phase; and
(iii)~the best \emph{fidelity} achieved during training, measured by the minimum sliced Wasserstein distance attained before memorization takes hold (i.e., for $\tau < \tau_{\mathrm{mem}}$).
The precise meaning of this last quantity will become clear after the discussion of the double descent phenomenon in Section~\ref{subsec:predictions-vs-reality}.

\subsection{Predictions versus reality: unexpected phenomena}
\label{subsec:predictions-vs-reality}

Before presenting the full hyperparameter sweeps, we describe the behavior of a single representative run (with default hyperparameters) and contrast it with our predictions based on the existing literature.

\begin{figure}[ht]
\begin{subfigure}{0.45\textwidth}
        \includegraphics[width=1\textwidth]{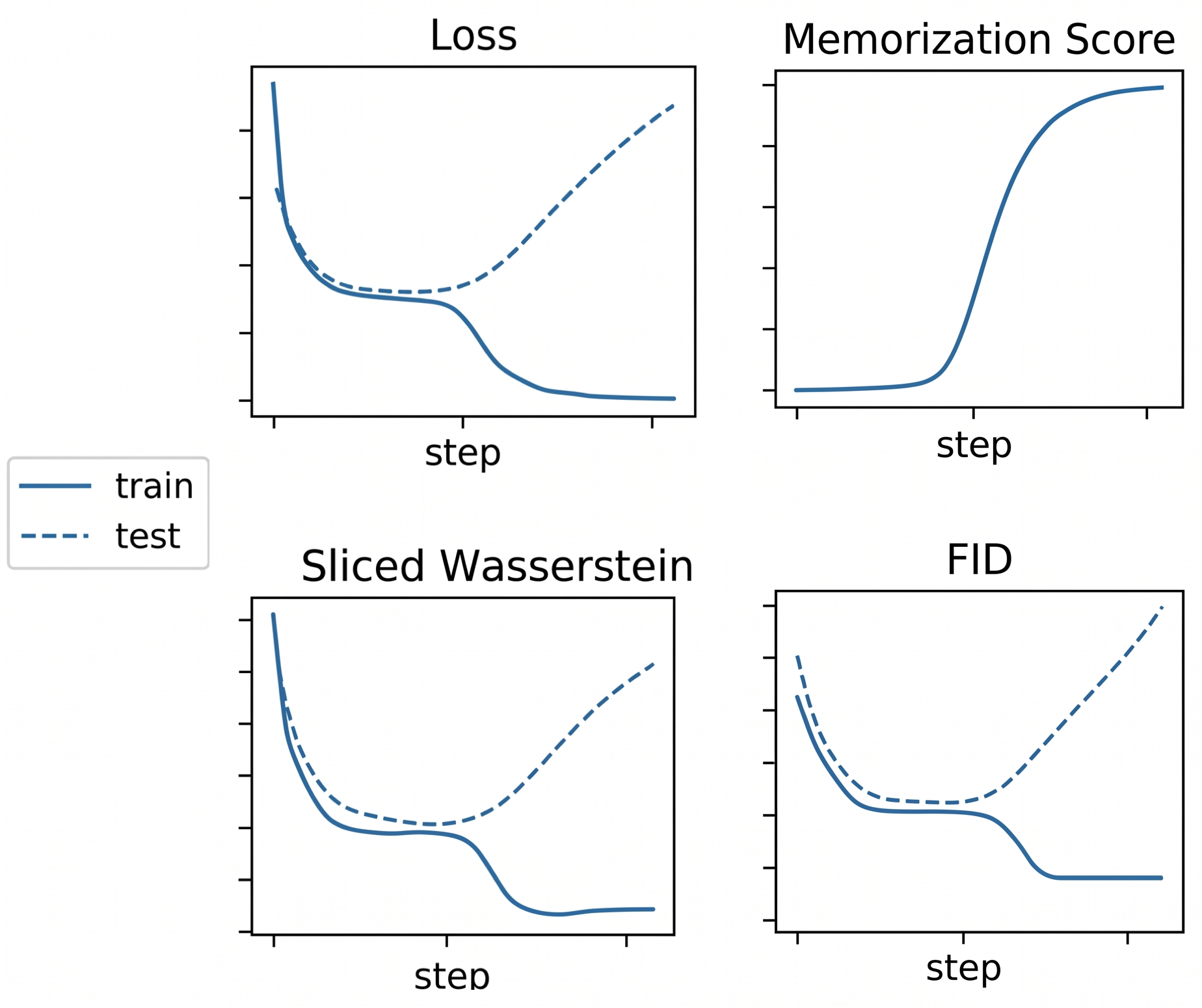}
        \caption{Our prediction before running the experiment.}
        \label{subfig:prediction}
    \end{subfigure}
    \hfill
    \begin{subfigure}{0.51\textwidth}
        \includegraphics[width=1\textwidth]{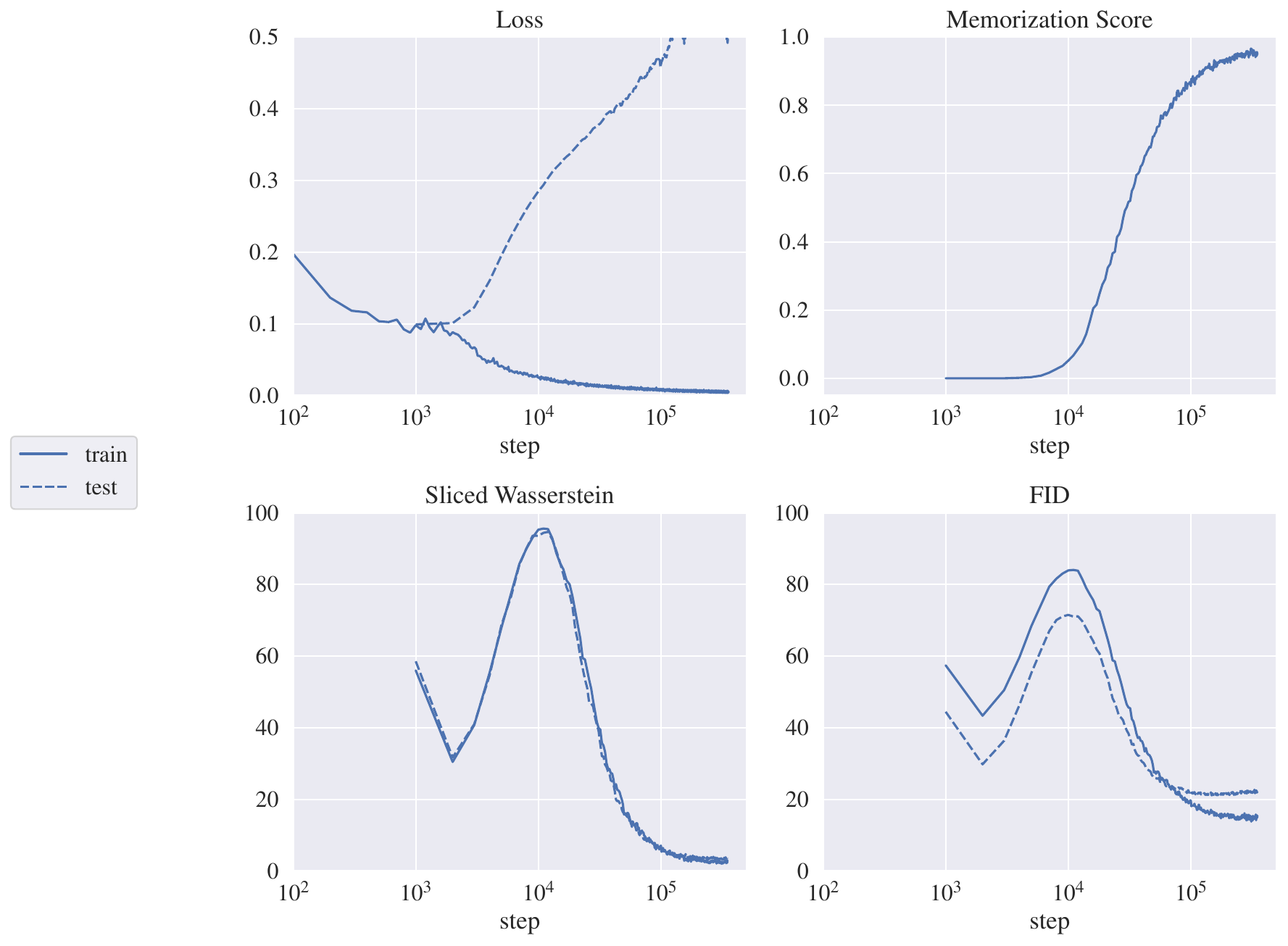}
        \caption{A representative run (default hyperparameters).}
        \label{subfig:representative-run}
    \end{subfigure}
\centering
\caption{Predicted versus actual evolution of training metrics.
\textbf{Left:} based on the literature (Section~\ref{subsec:implicit-reg}), we expected the train and test loss to decrease together until $\tau_{\mathrm{mem}}$, at which point the model overfits, and distributional distances to track the loss.
\textbf{Right:} actual experiment ($N=2048$, $P=92$M, $B=16$, $\eta=10^{-4}$).
The double descent in sliced Wasserstein distance and FID (bottom), present for \emph{both} train and test, is unexpected.}
\label{fig:representative-experiment}
\end{figure}

Informed by the literature described in Section~\ref{subsec:implicit-reg}, in particular~\citet{bonnaire2025diffusion} and~\citet{favero2025bigger}, we had predicted the behavior sketched in Figure~\ref{subfig:prediction}: the train and test losses would decrease together until~$\tau_{\mathrm{mem}}$, at which point the model would begin overfitting and the memorization score would rise.
Distances in distribution space (sliced Wasserstein, FID) would follow a similar trend to the loss, decreasing during the generalization phase and increasing once memorization begins.
The actual experiment (Figure~\ref{subfig:representative-run}) reveals two striking deviations from this prediction.

\paragraph{The train-test gap is uninformative about generative behavior.}
Throughout training, we observe little to no difference between the train and test distributional distances (sliced Wasserstein, FID).
Since the train and test splits are both drawn from the same distribution and are themselves close in distribution space, the triangle inequality implies that the distributional distance from generated samples to either split should be nearly identical.
The important implication is that \emph{none of the standard metrics we tracked correctly disentangles novelty from fidelity}:
distributional distances behave identically whether evaluated against the training or test set, even as the model transitions from producing novel images to producing copies.
The test loss can detect memorization but %
correlates poorly with visual quality \citep{theis2015note, blau2018perception}, and standard interventions such as classifier-free guidance \citep{ho2022classifier} improve perceptual quality while degrading the likelihood. This is the reason why practitioners choose FID over test loss to assess fidelity. 
This suggests that the classical supervised-learning notion of generalization, measured by a train-test performance gap either in score or distribution space, is insufficient for understanding generative model quality.
\paragraph{Double descent in distributional distance, for both train and test.}
Most unexpectedly, we observe a \emph{double descent} in distribution space (bottom panels of Figure~\ref{subfig:representative-run}).
Crucially, this double descent occurs for distances measured against \emph{both} the training and the test sets, ruling out explanations in terms of classical double descent or benign overfitting, which would manifest as a divergence between train and test metrics. Instead, this phenomenon appears driven purely by optimization dynamics: during an intermediate phase, the model reaches a state where it has low training loss yet generates a distribution that is far from the target. We note that \citet{yamaguchi2023limitation} made somewhat related empirical observations, linking them to dimensional reduction of the learned latent manifold. Experiments from \citet{bonnaire2025diffusion} on CelebA also feature the double descent for the FID (see Figure 2, left, particularly visible for $n=1024$), although authors do not comment on it.

One hypothesis is that different signal components are learned at different speeds: for instance, large-noise denoising (capturing high-level structure) before small-noise denoising (capturing fine details)~\citep{biroli2024dynamical,merger2025generalization,bardone2026theory}.
We tested this hypothesis by decomposing the train and test loss at various noise levels, but did not observe any meaningful pattern.
The mechanism underlying this double descent is thus an open question. We note that we did not observe it in similar experiments on ImageNet. We leave further investigation for future work.

\subsection{The roles of dataset size and model size}
\label{subsec:effect-dataset-model}
\begin{figure}[t]
    \centering
    \includegraphics[width=.8\linewidth]{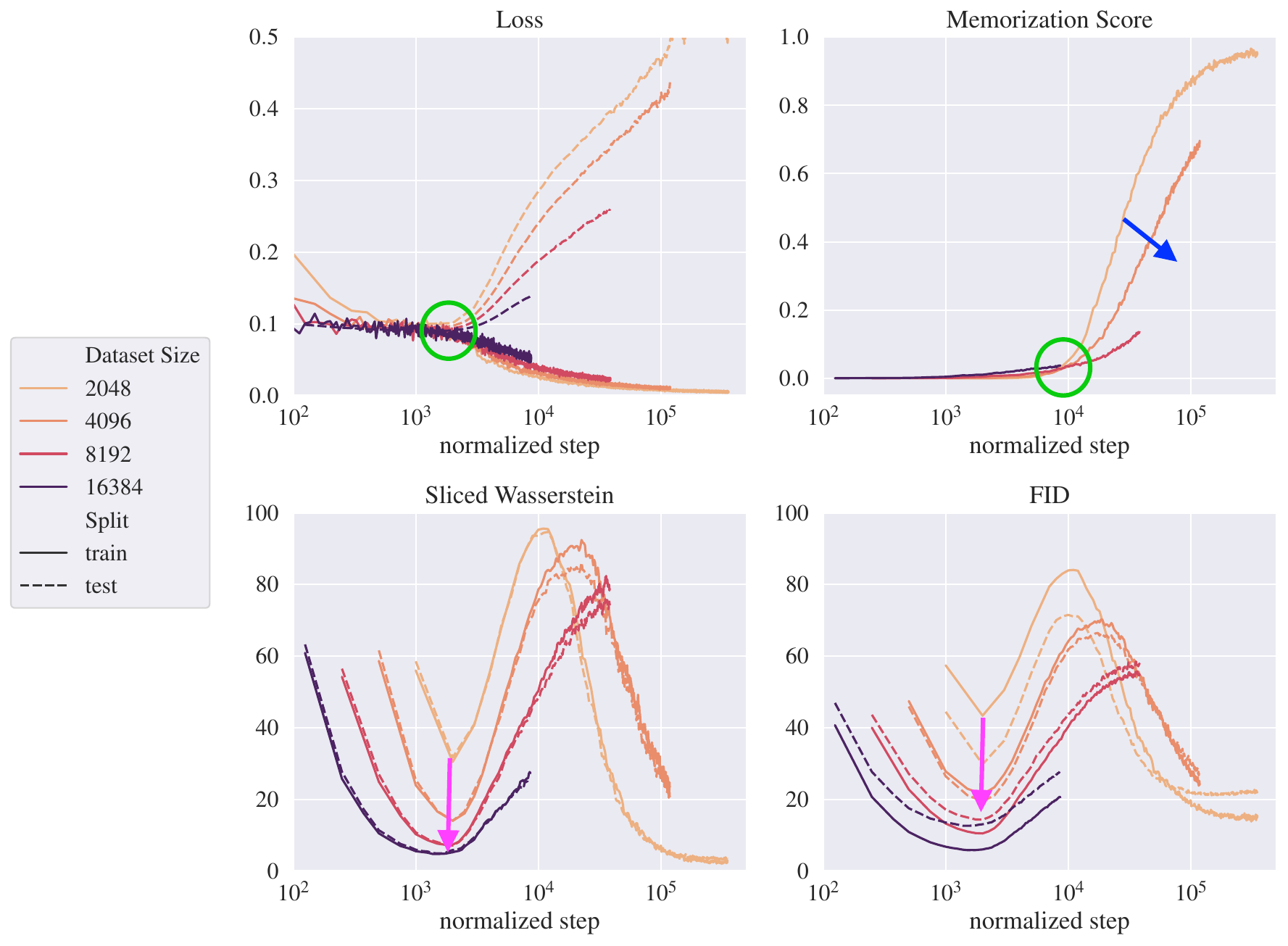}
    \caption{Effect of dataset size on training metrics (normalized step). Larger datasets delay memorization onset, reduce memorization rate, and improve peak fidelity.}
    \label{fig:ds_sweep}
\end{figure}
For clearer visualization across different settings, we plot metrics against the \emph{normalized training step}
\[
T \;=\; \tau \cdot \frac{N_{\min}}{N} \cdot \frac{B}{B_{\min}} \cdot \frac{\eta}{\eta_{\min}}\,,
\]
where $N_{\min}=2048$, $B_{\min}=16$, and $\eta_{\min}=10^{-5}$ are the smallest values in each sweep.
This rescaling accounts for the fact that larger datasets require more training steps, while larger batch sizes and learning rates process more information per step.
As we will see, the curves collapse under this rescaling with respect to $N$ and $B$ (confirming the linear scaling), though not for~$\eta$.

\paragraph{Effect of dataset size (Figure~\ref{fig:ds_sweep}).}
When increasing the dataset size, we observe three effects.
First, the normalized curves collapse before memorization, confirming that training time to memorization grows proportionally to the number of training samples. 
Second, the memorization rate (slope of the memorization score in the memorization phase) diminishes with larger~$N$, which is intuitive: it is harder to memorize a larger dataset.
Third, larger datasets lead to better fidelity, as measured by the minimum sliced Wasserstein distance achieved during training.

\begin{figure}[t]
    \centering
    \includegraphics[width=.8\linewidth]{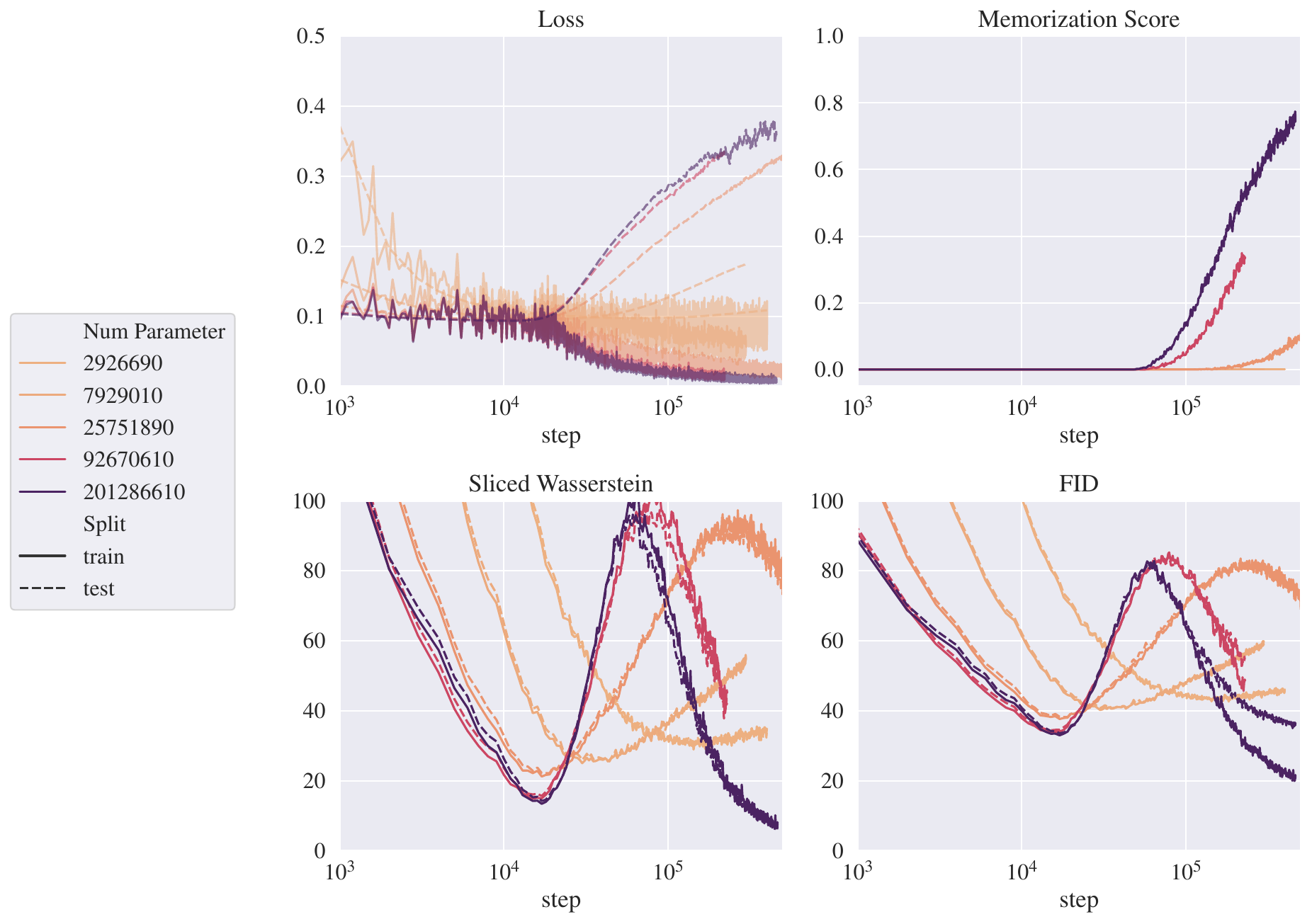}
    \caption{Effect of model size on training metrics. Larger models exhibit faster memorization onset, and peak fidelity is improved by increased capacity.}
    \label{fig:ms_sweep}
\end{figure}

\paragraph{Effect of model size (Figure~\ref{fig:ms_sweep}).}
When increasing the model size, we observe two effects: larger models begin memorizing sooner, which is reasonably intuitive by analogy with dataset size and there is an improvement in peak fidelity for larger models, both for train and test splits.

We highlight that the two sweeps play slightly different roles. Increasing the dataset size delays memorization and improve fidelity, whereas increasing the model size accelerates memorization while \emph{also} improving fidelity. This suggests that the classical statistical perspective, where generalization is governed by some ratio of model capacity to sample size, is insufficient to explain the phenomena observed here. The model's trajectory through parameter space, shaped by optimization dynamics, appears to play a role that cannot be reduced to a simple capacity argument.

\subsection{Optimization hyperparameters shape fidelity}
\label{subsec:effect-optim}

\begin{figure}[t]
    \centering
    \includegraphics[width=.8\linewidth]{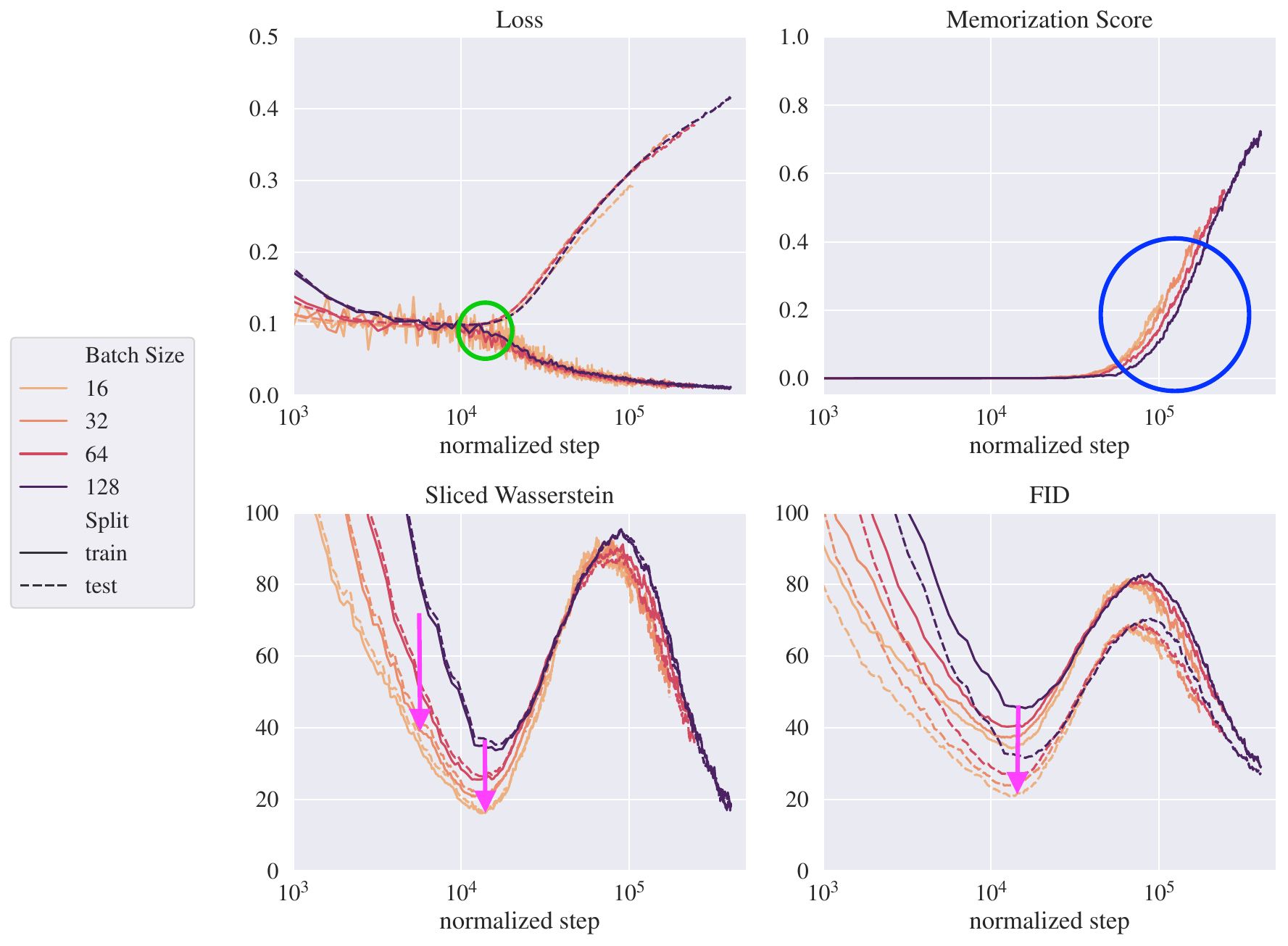}
    \caption{Effect of batch size on training metrics (normalized step). Smaller batch sizes improve peak fidelity while leaving the memorization onset and rate largely unchanged.}
    \label{fig:bs_sweep}
\end{figure}

\paragraph{Effect of batch size (Figure~\ref{fig:bs_sweep}).}
When increasing the batch size, we observe three effects.
First, the raw memorization onset time scales inversely with batch size, meaning that the normalized curves collapse.
Second, the memorization rate is essentially unaffected by batch size.
This is somewhat surprising, as one might expect that smaller batch sizes, which introduce more stochasticity, would slow down memorization.
Third, \emph{smaller batch sizes achieve better peak fidelity}.
This resonates with the minimum stability literature~\citep{mulayoff2024exact}, where smaller batches are known to bias SGD toward flatter minima.
However, the improved fidelity here manifests during training and not at convergence, a regime that is not directly addressed by the minimum stability framework.

\begin{figure}[t]
    \centering
    \includegraphics[width=.8\linewidth]{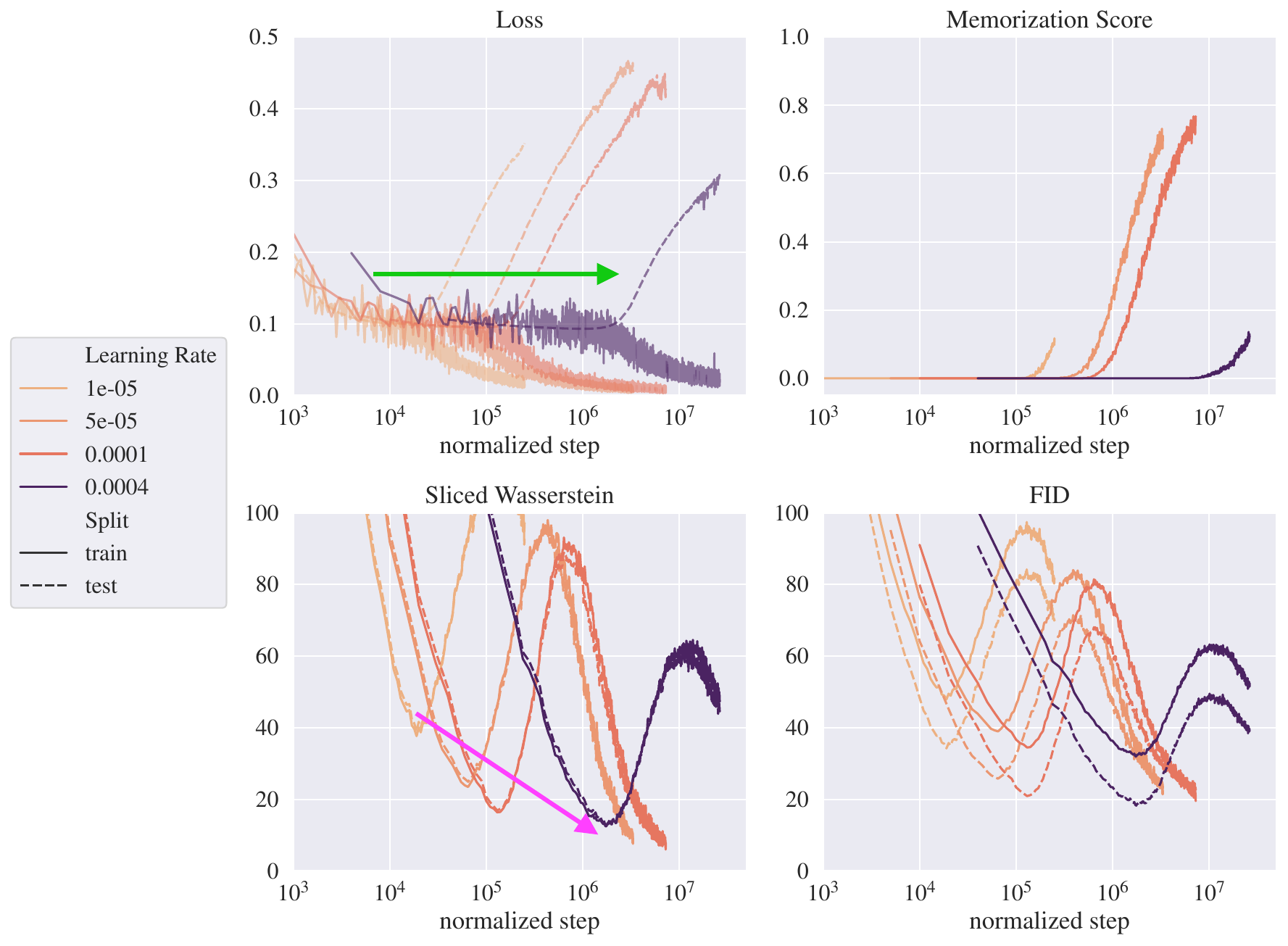}
    \caption{Effect of learning rate on training metrics (normalized step). Larger learning rates delay the normalized memorization onset and improve peak fidelity.}
    \label{fig:lr_sweep}
\end{figure}

\paragraph{Effect of learning rate (Figure~\ref{fig:lr_sweep}).}
When increasing the learning rate, we observe two notable effects.
First, the normalized training time \emph{increases} with larger learning rates.
This is consistent with the edge-of-stability phenomenon~\citep{cohen2021gradient}: beyond a certain learning rate, increasing~$\eta$ no longer proportionally accelerates training. Note that the largest learning rate we consider ($\eta = 4\cdot10^{-4}$) is close to the edge of stability in the sense that training diverges for larger values.
Second, larger learning rates lead to \emph{better peak fidelity}, consistent with the predictions of~\citet{wu2025taking}.
Again, however, this improvement manifests along the training trajectory rather than at convergence, suggesting that the implicit regularization from the learning rate operates differently from what the minimum stability framework predicts.

\section{Discussion: what we know and what remains open}
\label{sec:discussion}

We now distill the literature review of Section~\ref{sec:background} and the empirical findings of Section~\ref{sec:experiments} into what we believe are established facts and open questions.

\subsection{What is known}

\begin{enumerate}[topsep=0pt,itemsep=2pt,parsep=2pt,leftmargin=10pt]
    \item \textbf{The memorization transition is observable and systematic.}
    By carefully calibrating the dataset size, we can reliably observe the transition from a novelty phase to a memorization phase.
    This transition is gradual, with a well-defined onset~$\tau_{\mathrm{mem}}$ and a monotonically increasing memorization score thereafter.

    \item \textbf{Memorization onset scales linearly with dataset size.}
    This explains why memorization is rarely observed in practical large-scale training: the required training time simply exceeds any reasonable budget.
    In our view, this largely resolves the question of \emph{why} practical diffusion models do not memorize.
    The community should shift its focus from explaining the absence of memorization to understanding what the model learns during the pre-memorization phase and how fidelity emerges.

    \item \textbf{Model size and dataset size play different roles in the training dynamics.}
    The fidelity is improved both by increasing the dataset size and model size. On the contrary, memorization is delayed by increasing the dataset size but fastened by increasing the model size.

    \item \textbf{Optimization hyperparameters improve fidelity before the memorization transition.}
    Both smaller batch sizes and larger learning rates lead to better peak fidelity, as measured by the minimum distributional distance achieved before memorization.
    These effects echo predictions from the minimum stability literature but manifest along the training trajectory rather than at convergence. They do not have a strong impact after memorization starts.
\end{enumerate}

\subsection{Open questions}

\begin{enumerate}[topsep=0pt,itemsep=2pt,parsep=2pt,leftmargin=10pt]
    \item \textbf{What metrics adequately capture novelty, fidelity, and their interaction?}
    Standard distributional metrics (FID, sliced Wasserstein) fail to distinguish between a model that generalizes and one that memorizes, since both are evaluated against reference sets that are close in distribution space.
    Metrics such as precision and recall~\citep{kynkaanniemi2019improved} partially address this, but lack formal guarantees. A fundamental challenge is to develop metrics that are \emph{sensitive} to copy-vs-generalize distinction, \emph{robust} to the curse of dimensionality in image space, and ideally with known relationships to classical statistical divergences. More broadly, formalizing what it means for a generative model to "generalize" beyond its training data is a prerequisite for making the empirical study of memorization rigorous. 

    \item \textbf{How does implicit regularization operate along the training trajectory?}
    The minimum stability framework~\citep{mulayoff2024exact} characterizes the solution at convergence, but our findings suggest that the key regularization effects of learning rate and batch size manifest along the trajectory of optimization.
    This calls for a new theory of implicit regularization that describes the properties of intermediate iterates of SGD applied to the score matching objective, and characterizes the relationship between optimizer hyperparameters and the memorization-generalization balance at any given training step. 
    A plausible path towards this objective would be to apply the central flow theory of \citet{cohen2025understanding} to score matching.

    \item \textbf{How does the geometry of the data distribution shape generalization?}
    Understanding how properties of the data distribution, in particular its manifold structure and regularity, interact with the optimization dynamics and architecture to shape generalization is central to connecting theory and practice \citep{li2025scores,shen2026manifold}.
    Even in linear models, the structure of the data has been shown to interact in nontrivial ways with the training dynamics~\citep{merger2025generalization}.
    An interesting related direction is text-conditioned diffusion models, where the conditioning signal may provide a natural lever to disentangle novelty from fidelity, through assessing the fidelity of a conditional model conditioned on an unseen prompt.
\end{enumerate}

Beyond diffusion models, the theory of generalization in deep generative models requires a fundamentally different conceptual framework from the one that has been so successful for supervised learning.
In supervised learning, the training loss is a direct proxy for the quantity of interest (prediction accuracy), and the generalization gap is the natural object of study.
In generative modeling, the training loss (score matching) is at best an indirect proxy for sample quality, and the classical generalization gap, both in loss space and distribution space, carries little information about the model's generative behavior.

\section*{Acknowledgments} 

We thank Francis Bach, Quentin Berthet, G\'erard Biau, Linus Bleistein, Claire Boyer, Hugo Cui, L\'eo Dana, Alexandre Jiao, Zara Kadkhodaie, Claudia Merger, and Lenka Zdeborov\'a for fruitful discussions and suggestions.

\bibliographystyle{apalike}
\bibliography{ref}

\newpage
\appendix

\section{Technical details of the experiments}
\label{app:technical-details}

Full experimental details are given below. We also include details of parameters of the U-Net used.
\begin{table}[ht]
    \centering
    \begin{tabular}{lc}
    \toprule
    {\bf Name} & {\bf Value} \\
    \midrule
    Condition embedding dimension & 512\\
    Noise embedding dimension & 512\\
    Optimizer & Adam with standard hyperparameters \\
    EMA decay & $0.9999$ \\
    Hardware & 16 TPUv5 \\
    \midrule
    Noise schedule & Rectified Flow \\
    Number of sampling steps & $250$ \\
    CFG weight & 0.0 \\
    \midrule
    Number of base channels & 128 \\
    Attention Head Dimension & 64 \\
    Number of downsampling & 3 \\
    Channels multiplier & (1, 2, 4) \\
    Residual blocks per level & (3, 3, 3) \\
    \bottomrule
    \end{tabular}
    \vspace{.5em}
    \cprotect\caption{Common hyperparameters for training and sampling from diffusion models in all the experiments.}
    \label{tab:hyperparams}
\end{table}

\section{Sample examples}
\label{app:samples}

\begin{figure}[ht]
\begin{subfigure}{0.3\textwidth}
        \includegraphics[width=0.35\textwidth]{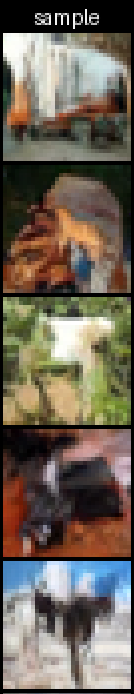}
        \caption{$N=2048$, generalizing model.}
    \end{subfigure}
    \hfill
    \begin{subfigure}{0.3\textwidth}
        \includegraphics[width=0.35\textwidth]{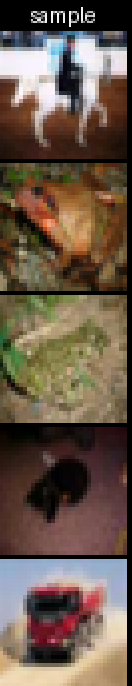}
        \caption{$N=2048$, memorizing model.}
    \end{subfigure}
    \hfill
    \begin{subfigure}{0.3\textwidth}
        \includegraphics[width=0.35\textwidth]{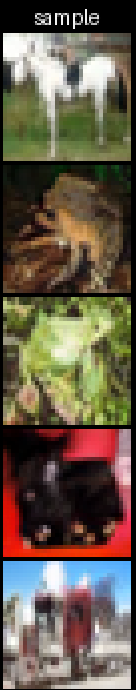}
        \caption{$N=16{,}384$, generalizing model.}
    \end{subfigure}
\centering
\caption{Generated samples at different training stages and dataset sizes, with the same randomness for each row.
\textbf{(a)}~During the generalization phase with $N=2048$, the model produces diverse images, though quality is limited by the small dataset.
\textbf{(b)}~After memorization with $N=2048$, the model generates near-copies of training data.
\textbf{(c)}~During the generalization phase with $N=16{,}384$, the model produces more diverse, higher-quality samples, illustrating the benefit of larger training sets.}
\label{fig:sample-examples}
\end{figure}

\end{document}